\documentclass[runningheads]{llncs}
\usepackage{graphicx}

\usepackage{tikz}
\usepackage{comment}
\usepackage{amsmath,amssymb} 
\usepackage{color}
\usepackage{tikz}
\usepackage{comment}
\usepackage{amsmath,amssymb} 
\usepackage{color}
\usepackage{amsmath}
\usepackage{graphicx}
\usepackage{amsmath}
\usepackage{amssymb}
\usepackage{booktabs}
\usepackage{multirow}
\usepackage{multicol}
\usepackage{epsfig}
\usepackage{graphicx}
\usepackage{amsmath}
\usepackage{amssymb}
\usepackage{bm}
\usepackage{booktabs}
\usepackage{multirow}
\usepackage{threeparttable}
\usepackage{soul}
\usepackage{wrapfig}
\usepackage{hhline}
\usepackage{breakcites}

\usepackage[accsupp]{axessibility}  

\usepackage[section]{placeins}
\usepackage{float}

\usepackage[capitalize]{cleveref}
\crefname{section}{Sec.}{Secs.}
\Crefname{section}{Section}{Sections}
\Crefname{table}{Table}{Tables}
\crefname{table}{Tab.}{Tabs.}

\newcommand{\eg}{\textit{e.g.}}
\newcommand{\ie}{\textit{i.e.}}
\newcommand{\etal}{\emph{et al.}}

\begin{document}
\pagestyle{headings}
\mainmatter
\def\ECCVSubNumber{283}  

\title{SiRi: A Simple Selective Retraining Mechanism \\ for Transformer-based Visual Grounding}

\titlerunning{SiRi}
%
\author{Mengxue Qu\inst{1, 2}\thanks{Work done during an internship at  JD Explore Academy. } \and
Yu Wu\inst{3} \and
Wu Liu\inst{4} \and
Qiqi Gong\inst{1,2} \and
Xiaodan Liang\inst{5} \and
\\Olga Russakovsky\inst{3}\and
Yao Zhao\inst{1,2} \and
Yunchao Wei\inst{1,2}}

\authorrunning{M. Qu et al.}

\institute{ $^1$Institute of Information Science, Beijing Jiaotong University \\
$^2$Beijing Key Laboratory of Advanced Information Science and Network Technology \\
$^3$Princeton University   \quad
$^4$JD Explore Academy  \quad
$^5$Sun Yat-sen University \\
\email{qumengxue@bjtu.edu.cn}, \email{yuwu@princeton.edu}, \email{wychao1987@gmail.com}
}

\maketitle
\begin{abstract}

In this paper, we investigate how to achieve better visual grounding with modern vision-language transformers, and propose a simple yet powerful \textbf{S}elect\textbf{i}ve \textbf{R}etra\textbf{i}ning (SiRi) mechanism for this challenging task. Particularly, SiRi conveys a significant principle to the research of visual grounding, \ie, a better initialized vision-language encoder would help the model converge to a better local minimum, advancing the performance accordingly.
In specific, we continually update the parameters of the encoder as the training goes on, while periodically re-initialize rest of the parameters to compel the model to be better optimized based on an enhanced encoder. SiRi can significantly outperform previous approaches on three popular benchmarks. Specifically, our method achieves 83.04\% Top1 accuracy on RefCOCO+ \emph{testA}, outperforming the state-of-the-art approaches (training from scratch) by more than 10.21\%. Additionally, we reveal that SiRi performs surprisingly superior even with limited training data. We also extend it to transformer-based visual grounding models and other vision-language tasks to verify the validity.
Code is available at \url{https://github.com/qumengxue/siri-vg.git}.

\keywords{Visual grounding, Transformer, Generalization}

\end{abstract}

\section{Introduction}
\label{sec:intro}

Visual grounding~\cite{RefCOCO_and_REFCOCO+,RefCOCOg}, also known as Referring Expression Comprehension (REC), aims to predict the location of a region referred to by the language expression in an image. Previous solutions can be roughly divided into two-stage methods~\cite{19LearningToCompose,CMN,28LearnigToAssemble,46Learning2Branch,48neighbourhood,52dynamic,59mattnet,63grounding,68parallel} and one-stage methods~\cite{9real,27real,42zero,55improving,56fast}. The two-stage methods start with the process of generating region proposals via object detectors~\cite{FastRCNN} and then learning to identify the expected object from hundreds of candidates. On the other hand, the one-stage methods perform the grounding in an end-to-end manner, and often with inferior performances. However, the performance of these models is significantly limited due to the huge semantic gap between diverse referring descriptions and various visual appearances. The reason is that visual grounding needs to consider many open or fine-grained (\eg, {girl}, {boy}, {child}) categories, which is significantly different from the common vision tasks (\eg, classification, detection, and segmentation) where each image or individual object has a clear class label. Therefore, due to the diversity of descriptions in the human world, the model may easily overfit the descriptions in \textit{train} while hard to correctly understand the referring expressions in \textit{val} and \textit{test} when the training data is insufficient.

Recently, many researchers focus on using the attention mechanism in Transformer for Vision-Language (V-L) modeling \cite{VLBERT,ViLBERT,TransVG,MDETR}. With both visual and linguistic elements as the inputs, the Transformer encoder can perceive multi-modal data and thoroughly model the visual-linguistic relationship. Although these Transformer-based methods have achieved great success in vision-language modeling, they heavily rely on pre-training with extra large-scale vision-language data pairs to improve the generalization ability of the encoder and relieve the over-fitting issue, accordingly. 

\begin{wrapfigure}[20]{r}{6cm}
\centering
\includegraphics[width=1.0\linewidth]{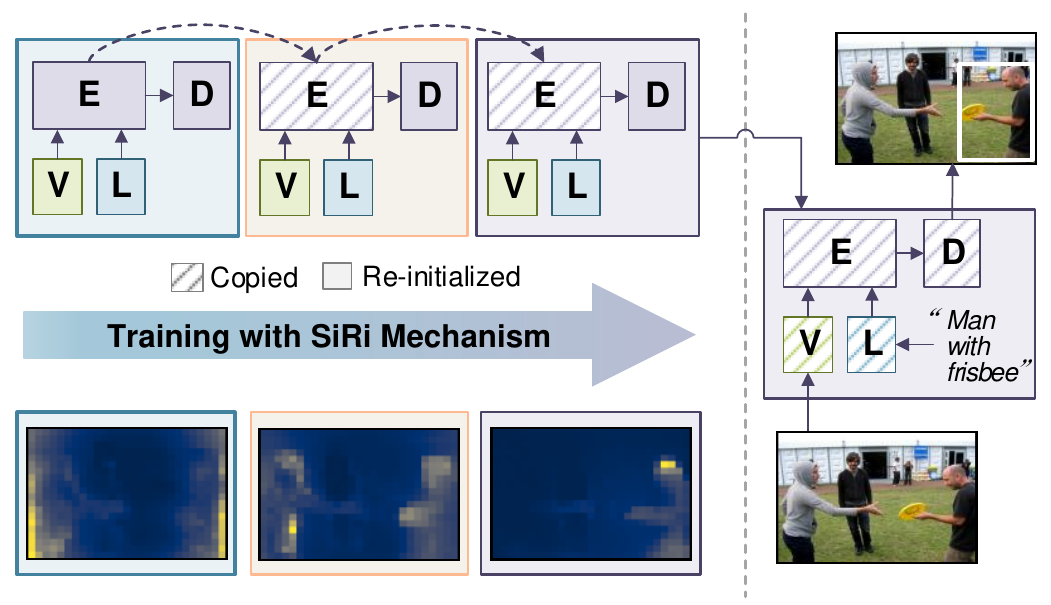}
\caption{The sketch of our SiRi mechanism of three retraining periods. ``V'': Visual Backbone, ``L": Language Backbone, ``E'': Visual-Language Transformer Encoder, ``D": Transformer Decoder. 
   The right part shows that we only take the last retrained model for the final test. Best viewed in color.}
\label{fig:onecol}
\end{wrapfigure}
However, without large-scale data pre-training, the model shows significant performance degradation on visual grounding tasks. 
We observe that the relationship between the given expression and the image perceived by the Transformer encoder leaves much to be desired based on the poor V-L interaction attention map in Fig.~\ref{fig:onecol}.
The reason may be that the Transformer encoder, started with randomly initialized parameters, may easily over-fit a small number of training pairs and make the model be trapped into a poor local minimum. With such an observation, we raise the question of \emph{whether the V-L model will converge to a better local minimum by equipping the Transformer encoder with better-initialized parameters?}

To answer the above question, in this paper, we investigate a new training mechanism to improve the Transformer encoder, named \textbf{S}elect\textbf{i}ve \textbf{R}etra\textbf{i}ning (SiRi), which repeatedly reactivates the learning of the encoder in the process of continuous retraining and progressively provide better-initialized parameters for the encoder in the next stage.
Specifically, while we \textit{continually update} parameters of the encoder as the training goes on, we \textit{periodically re-initialize} all the other modules (\eg, vision/language backbones and the Transformer decoder). In this way, the SiRi promotes the encoder to continually learn better vision-language relationships by periodically getting out of the sub-optimal saddle point. Fig.~\ref{fig:onecol} shows the sketch of SiRi and the visualization of the encoder's attention weight after each retraining period, where we can clearly see the progress of the encoder in multi-modal modeling.

We conduct extensive experiments to validate the effectiveness of our method. 
With the proposed SiRi mechanism, our model remarkably outperforms previous approaches on three popular benchmarks. Particularly, we achieve 83.04\% at top-1 accuracy on RefCOCO+ \emph{testA}~\cite{RefCOCO_and_REFCOCO+}, outperforming the state-of-the-art approaches by more than 10.21\%. 

\begin{figure}[t]
  \centering
    \includegraphics[width=1.00\linewidth]{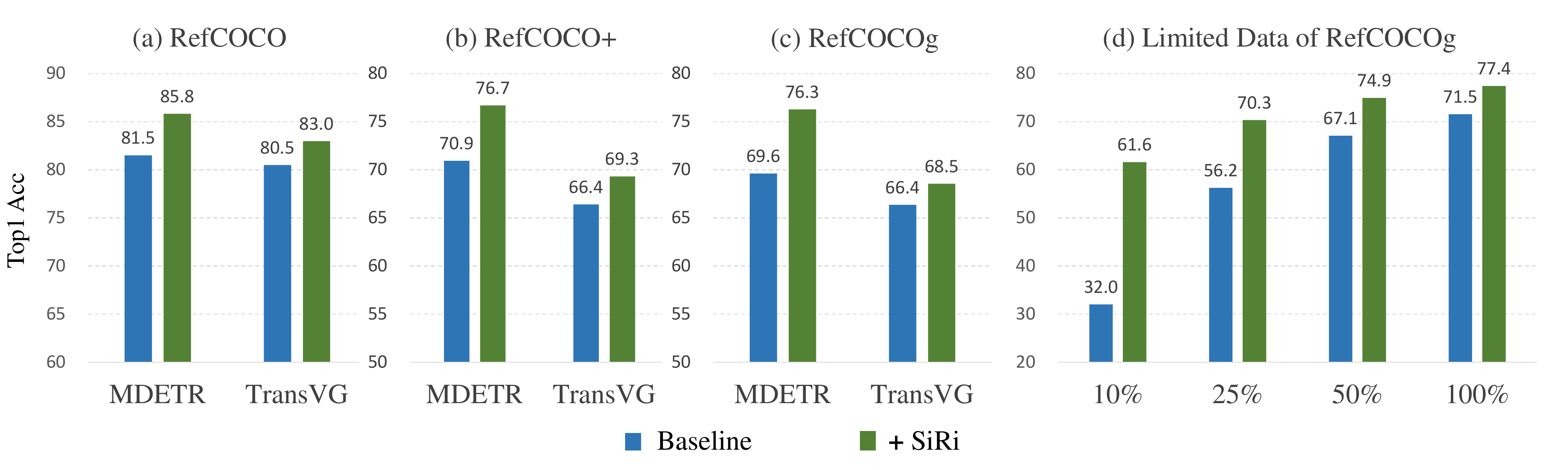}

  \caption{(a)-(c) illustrates the performance enhancement of SiRi on MDETR~\cite{MDETR} and TransVG~\cite{TransVG}. We test on three popular visual grounding datasets RefCOCO, RefCOCO+, RefCOCOg. (d) shows that when training with 10\%, 25\%, 50\%, 100\% \emph{training} data, the top1 accuracy improvement of SiRi on the RefCOCOg \emph{validation} set.
  }
  \label{fig:bar_chart}
\end{figure}

More importantly, we further observe that the SiRi mechanism helps model generalize well to small-scale training data as shown in Fig.~\ref{fig:bar_chart} (d). To be specific, our model with a quarter of training data outperforms previous state-of-the-art methods (with full training data) by 1.65\% on the RefCOCOg \textit{val} set. With even less training data (\eg, only 10\%), we almost double the accuracy (61.58\% \textit{versus} 32.00\%) compared to the baseline. Additionally, we complement more extensibility studies in other visual grounding model and other V-L tasks related to visual grounding. We found SiRi can further improve the top-1 accuracy by an average of 2\% in TransVG~\cite{TransVG}, which is also a Transformer-based visual grounding model. We visualize the improvement of different model with SiRi on three datasets in Fig.~\ref{fig:bar_chart} (a) - (c). In other V-L tasks, including referring expression segmentation, phrase grounding, and visual question answering tasks, we can also improve the baseline using the SiRi mechanism. 

\section{Related Work}
\label{sec:formatting}

\subsection{Visual Grounding}
Existing methods for Visual Grounding based on CNN can be roughly divided into two categories, namely two-stage methods and one-stage methods.  

\noindent \textbf{Two-stage methods}~\cite{19LearningToCompose,CMN,liang2021clawcranenet,liang2021rethinking,28LearnigToAssemble,46Learning2Branch,48neighbourhood,wu2022snoc,52dynamic,59mattnet,63grounding,68parallel} typically utilize an object detector to generate region proposals in the first stage, and then find the best matched region-text pair. The object-text pair matching is commonly used in visual grounding task and other V-L tasks, \eg, retrieval tasks~\cite{HGA}. MattNet~\cite{59mattnet} takes a modular approach to progressively understand and unify visual and linguistic semantic information in terms of attributes, relationships, and location. Additionally, some approaches further enhance the modeling ability of multi-modal relations using graph structures~\cite{48neighbourhood,52dynamic,53graph}, multi-modal tree structures~\cite{28LearnigToAssemble}. 

\noindent
\textbf{One-stage methods}~\cite{9real,27real,42zero,55improving,56fast} avoid being constrained by the quality of the proposal by directly fusing visual and linguistic features.
FAOA~\cite{56fast} represents the text input with a language vector and leverages it into the YOLOv3 detector~\cite{40yolov3} to align the referred instance. RCCF~\cite{27real} regards the visual grounding problem as a correlation filtering process~\cite{4visual,17high}, and the peak value in the correlation heatmap is selected as the center of target objects. In ReSC~\cite{55improving}, the limitation of FAOA~\cite{56fast} on grounding complex queries is broken through with a recursive sub-query construction module.

In the previous CNN-based visual grounding model, the V-L fusion is performed throughout the decoding process, which is weak interpretability and performance compared to the V-L fusion module in Transformer-based model. Therefore, we adopt Transformer-based model for better V-L interaction.

\subsection{Transformer-based Methods in REC}
Recently, Transformer~\cite{45Transformer} has been widely used to address the multi-modal semantic alignment problem. 
However, Transformer is data-hungry and thus usually needs additional large-scale pretraining.
Motivated by the excellent performance of BERT~\cite{12bert}, some researchers~\cite{VLBERT,UNITER,ViLBERT,ERNIE-ViL,VILLA,ViLT,LXMERT} construct similar structures and propose multi-modal pre-training for Visual-Language Pre-training (VLP) tasks. These approaches introduce pretext tasks for better interaction of vision and language, \eg, masked language modeling~\cite{ViLBERT,VLBERT}, image-text matching~\cite{ViLT}. 
However, these VLP methods usually require pre-training with large-scale data and fine-tuning on downstream tasks to achieve good results.
Recently, TransVG~\cite{TransVG} study the Transformer-based framework without pretraining. 
Without extracting region proposals in advance, TransVG directly regresses bounding box coordinates and predicts the referring objects. 

These works have validated the effectiveness of Transformer for multimodal modeling. However, most of them require large-scale data to pretrain a Transformer-based model. 
Differently, in this work, we focus on exploring a way to train better encoders \textit{without} large-scale pretraining.

\subsection{Re-training}

Some early works avoid getting trapped in a local minimum by introducing randomness. For example, ensemble learning~\cite{NN_Ensembles_1,NN_Ensembles_2} introduces randomness by retraining the model with different random initialized parameters to converge to different local minimums. %
Due to these studies requiring an overwhelming cost, a number of retraining methods, \eg, Dropout~\cite{dropout}, Distillation~\cite{distilling}, are proposed to reduce the cost of retraining in ensemble learning. More recently, Snapshot Ensemble~\cite{Snapshot} proposes to retrain the same model to access multiple local minimums by the cyclic learning rate. Similarly, the cyclic learning rate is used in the retraining process to detect noisy labels in O2U-Net~\cite{o2u-net}. However, Transformer~\cite{45Transformer} is very sensitive to the learning rate and sometimes requires a warm-up or inverse square root learning rate, which makes the cyclic learning rate~\cite{cycle_lr} inapplicable. The proposed weight initialization scheme T-Fixup in~\cite{Transformer_better_initial} enables Transformer training without warmup or layer normalization. Han~\etal~\cite{DSD} proposes DSD retraining mechanism with reference to the model pruning, which avoids over-fitting caused by over-capturing of noisy data.

The SiRi mechanism proposed in this paper is somehow similar to the above methods but SiRi is designed for the V-L fusion module in V-L tasks. The main motivation of re-training in this paper is to provide the V-L fusion Transformer with better-initialized parameters.

\section{Method}
\label{sec:method}

In this section, we first briefly review the basic visual grounding architecture adopted by this work in Sec.~\ref{subsec:base arch.}. Then we elaborate on our proposed SiRi mechanism in Sec.~\ref{subsec:SiRi} and the Multi-task SiRi in Sec.~\ref{subsec:dual decoder}.


\subsection{Base Architecture}
\label{subsec:base arch.}
We follow the state-of-the-art model MDETR~\cite{MDETR} as our base architecture, which consists of four main modules: (1) Visual Backbone; (2) Language Backbone; (3) Visual-Language Transformer Encoder; (4) Transformer Decoder Module. 

\noindent \textbf{Visual Backbone $\mathcal{V}$} \& \textbf{Language Backbone $\mathcal{L}$.}
We adopt the convolutional backbone ResNet-101~\cite{ResNet} to obtain the visual representation for an input image ${\mathbf{I}}$.
In previous work MDETR~\cite{MDETR}, they only take the output of the last CNN stage as visual features.
Differently, we believe the features of shallow stages (\eg, the third stage in ResNet-101) benefit localizing objects if the sentence contains a detailed low-level description such as color.
Therefore, we take the output of the third stage of ResNet-101 and transform it with two dilated convolution layers.
Then we add the adjusted dimensionality low-level feature together using the final-stage output of ResNet-101 as the final visual representations.
Then we encode referring expressions with the pretrained language model RoBERTa~\cite{RoBERTa}.

\begin{figure*}[!t]
\begin{center}
\includegraphics[width=1.00\linewidth]{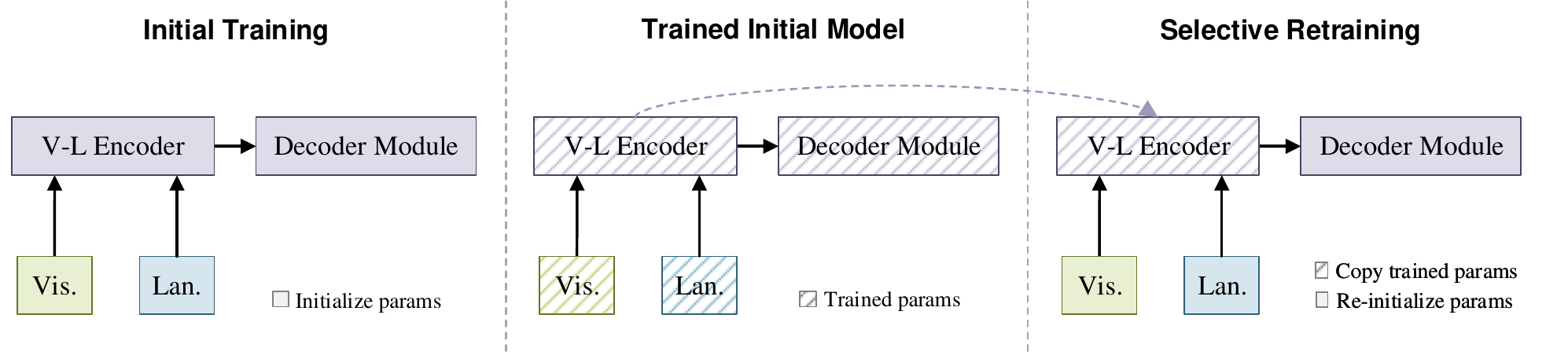}
   \caption{The training process of our SiRi mechanism. The parameters of the module with solid color background are initialized as the original rules, while those with slash background are trained. The base architecture contains four main modules: (1) ``Vis.": Visual Backbone; (2) ``Lan.": Language Backbone; (3) ``V-L Encoder": Visual-Language Transformer Encoder; (4) ``Decoder Module": Transformer Decoder Module.}

\label{fig:arch}
\end{center}
\end{figure*}

\noindent \textbf{Visual-Language Transformer Encoder $\mathcal{E}$.} 
We use a Transformer~\cite{45Transformer} as the encoder for vision-language interaction, where the model performs the cross-modal fusion and association.
To do so, we flatten the visual features and add 2-D positional embeddings to conserve spatial information.
After that, we project both the flattened visual features and text features into a shared embedding space and then concatenate them into a single sequence of image and text features.
The sequence is then input to the cross encoder Transformer for further visual-language interaction.

\noindent \textbf{Transformer Decoder $\mathcal{D}$.} Following DETR~\cite{DETR}, we use a Transformer decoder to predict the target bounding boxes.
The decoder takes as input a set of learnable object queries, cross-attends to the encoder output and predicts embeddings for each query. 
After that, we decode the embeddings into box coordinates and class labels by the regression and classification heads.
Considering that the number of relevant referred targets is fewer than the total number of objects of an image, we limit the decoder to have 16 query inputs only. 
Considering there is only sentence-level correspondence in visual grounding, we remove box-token contrastive alignment loss~\cite{MDETR}.
Accordingly, we also reduce the length of the soft tokens to 2, standing for whether the object box belongs to the expression.

%
\subsection{SiRi: Selective Retraining Mechanism}
\label{subsec:SiRi}

\begin{wrapfigure}[15]{r}{6cm}
\centering
\includegraphics[width=0.9\linewidth]{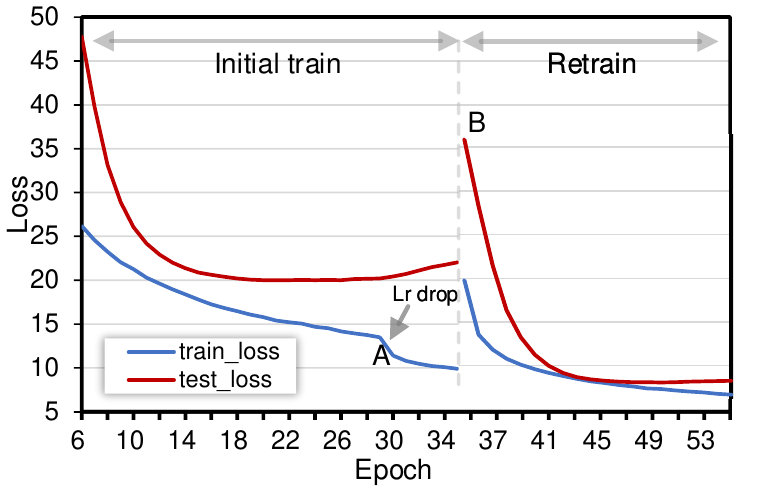}

\caption{The train and test loss curves in \textit{Initial train} stage and \textit{Retrain} stage. }
\label{fig:curve}
\end{wrapfigure}

The transformer model may easily get over-fitted without large-scaled pretraining.
As shown in Fig.~\ref{fig:curve}, the test loss increases even though the training loss still declines after point A of the initial training stage.
Simply having more training iterations would not further improve the test performance.

Motivated by our hypothesis that a V-L model may converge to a better local minimum by equipping the Transformer encoder with better initialized parameters, we design the Selective Retraining (SiRi) mechanism.
After the initial training, we continually update the parameters of the encoder as the training goes on, while periodically re-initializing the parameters of the decoder to compel the model to be better optimized based on an enhanced encoder. By applying our SiRi mechanism at point B in Fig.~\ref{fig:curve}, both {training loss} and {test loss} further decline, thus we obtain better optimization results (lower test loss). 
To be specific, our Selective Retraining Mechanism is set up as follows.

\textbf{Initial Training.} 
We initialize the visual Backbone $\mathcal{V}$  and the language Backbone $\mathcal{L}$ using the ResNet-101~\cite{ResNet} model pre-trained from ImageNet~\cite{ImageNet} and the RoBERTa model pre-trained from language corpus datasets, respectively.

The rest of our model (\eg, Transformer encoder and decoder) are randomly initialized using the Xavier initialization~\cite{Xavier}.
We denote the initialized parameters of the Visual Backbone together with the visual linear projection layer as $ {\mathcal{V}_{0}}$, and Language Backbone together with the corresponding linear projection layer as $ {\mathcal{L}_{0}}$.
Similarly, the model weights of Transformer Encoder and Transformer Decoder are denoted as $ \mathcal{E}_{0}$ and $ {\mathcal{D}_{0}}$, respectively.
We then train the model using a combination of the object coordinates regression losses (L1 \& GIoU) and soft-token prediction loss (cross-entropy loss) while keeping the learning rate unchanged.
The model training stops when the validation performance stays stable. 
We denote the trained model weights to be $ {\mathcal{V}_{0}^{'} },  {\mathcal{L}_{0}^{'} },  {\mathcal{E}_{0}^{'} },  {\mathcal{D}_{0}^{'} }$ after the initial training.

\textbf{Selective Retraining.} 
To further improve the encoder with better vision-language understanding, we continually train the encoder after the initial training, while \textit{re-initialize} the other modules to avoid getting stuck in local minimums. We show the pipeline of SiRi in Fig.~\ref{fig:arch}.
Specifically, for the $t$-th round of the selective retraining, we only keep the encoder $\mathcal{E}_{t}$ to be up-to-date, \ie, $\mathcal{E}_{t} \gets \mathcal{E}_{t-1}^{'}$, where $\mathcal{E}_{t-1}^{'}$ is the previous trained encoder from $t-1$ round.
As for other modules including the decoder $\mathcal{D}_{t} $, the visual backbone $\mathcal{V}_{t}$, and the language backbone $\mathcal{L}$, we drop the trained weights and re-initialize them using their original initialization at the initial training stage, \ie, either initializing from the pre-trained weights (\eg, $\mathcal{V}_0$ and $\mathcal{L}_0$), or random initialization (\eg, the decoder $D$).
We then re-train the whole model using the same learning rate until it converges.

\subsection{Multi-task SiRi}
\label{subsec:dual decoder}

As a common practice for transformer models, multi-task learning usually benefits the model optimization and thus alleviates over-fitting issues.
Therefore, we further extend SiRi to a multi-task version by incorporating an auxiliary decoder.
Specifically, we use two diverse decoders to generate predictions based on the same encoder output and then optimize the encoder using the two decoder losses.

To ensure the two decoders are different from each other, we design two different object queries (positional embeddings) for decoders.
Previous DETR~\cite{DETR} uses \textit{learnable} positional embeddings as the object query to attend to the encoder output.
Differently, we adopt a {\textit{constant}} positional encoding sequence, \ie, the sine-cosine position encoding function, to generate the object queries for the other decoder. 
The two decoders take different queries to attend to the same encoder output, which would urge the encoder to be more robust in vision-language interaction.
The details are shown in Appendix~\ref{subsec:dual decoder details}.

\section{Experiments}

\subsection{Datasets}
\label{sec:data}

\textbf{RefCOCO/RefCOCO+ } are proposed in ~\cite{RefCOCO_and_REFCOCO+}. There are 19,994 images in RefCOCO  with 142,209 refer expressions for 50,000 objects. Similarly, 19,992 images are included in RefCOCO+ which contains 141,564 expressions for 49,856 objects. In these datasets, each image contains two or more objects from the same category. In RefCOCO+ dataset, positional words are not allowed in the referring expression, which is a pure dataset with appearance-based referring expression, whereas RefCOCO imposes no restriction on the phrase.
In addition to the training set and validation set, the test set for RefCOCO/RefCOCO+ is divided into a \textit{testA} set (containing several people in an image) and a \textit{testB} set (containing multiple instances of other objects in an image).

\textbf{RefCOCOg}~\cite{RefCOCOg} contains 26,711 images with 85,474 referring expressions for 54,822 objects, and each image usually contains 2-4 objects of the same category. The length of referring expressions in this dataset is almost twice as long as those in RefCOCO and RefCOCO+. %

\begin{table}[!t]
\centering
\small
\caption{Comparisons with state-of-the-art methods on RefCOCO~\cite{RefCOCO_and_REFCOCO+}, RefCOCO+~\cite{RefCOCO_and_REFCOCO+}, and RefCOCOg~\cite{RefCOCOg} in terms of top-1 accuracy. We also report official MDETR implementation~\cite{MDETR} without pretraining (denoted as MDETR w/o pretrain) and our improved MDETR implementation (see Sec.~\ref{subsec:base arch.}) (denoted as MDETR*).  ``MT SiRi" means ``Multi-task SiRi". }
\scalebox{0.78}{
\begin{tabular}{c|c|c|ccc|ccc|cc} 
\toprule
\multirow{2}{*}{Method}                          & \multirow{2}{*}{Venue} & \multirow{2}{*}{Visual backbone} & \multicolumn{3}{c|}{RefCOCO}                                                       & \multicolumn{3}{c|}{RefCOCO+}                                                      & \multicolumn{2}{c}{RefCOCOg}                           \\
                                                 &                        &                                  & val                       & testA                     & testB                      & val                       & testA                     & testB                      & val                       & test                       \\ 
\hline\hline
\textit{CNN-based:}                              & \multicolumn{1}{l|}{}  & \multicolumn{1}{l|}{}            & \multicolumn{1}{l}{}      & \multicolumn{1}{l}{}      & \multicolumn{1}{l|}{}      & \multicolumn{1}{l}{}      & \multicolumn{1}{l}{}      & \multicolumn{1}{l|}{}      & \multicolumn{1}{l}{}      & \multicolumn{1}{l}{}       \\
CMN~\cite{CMN}                                             & CVPR'17                & VGG16~\cite{VGG}                           & -                         & 71.03                     & 65.77                      & -                         & 54.32                     & 47.76                      & -                         & -                          \\
MAttNet~\cite{59mattnet}                                         & CVPR'18                & ResNet-101~\cite{ResNet}                       & 76.65                     & 81.14                     & 69.99                      & 65.33                     & 71.62                     & 56.02                      & 66.58                     & 67.27                      \\
RvG-Tree~\cite{19LearningToCompose}                                        & TPAMI'19               & ResNet-101                       & 75.06                     & 78.61                     & 69.85                      & 63.51                     & 67.45                     & 56.66                      & 66.95                     & 66.51                      \\
NMTree~\cite{28LearnigToAssemble}                                          & ICCV'19                & ResNet-101                       & 76.41                     & 81.21                     & 70.09                      & 66.46                     & 76.02                     & 57.52                      & 65.87                     & 66.44                      \\
FAOA~\cite{56fast}                                            & ICCV'19                & DarkNet-53~\cite{40yolov3}                      & 72.54                     & 74.35                     & 68.50                      & 56.81                     & 60.23                     & 49.60                      & 61.33                     & 60.36                      \\
RCCF~\cite{27real}                                            & CVPR'20        & DLA-34~\cite{DLA}                          & -                  & 81.06                     & 71.85                      & -                         & 70.35                     & 56.32                      & -                         & 65.73                      \\
MCN~\cite{MCN}                                             & CVPR'20                & DarkNet-53                       & 80.08                     & 82.29                     & 74.98                      & 67.16                     & 72.86                     & 57.31                      & 66.46                     & 66.01                      \\
ReSC-Large~\cite{55improving}                                      & ECCV'20                & DarkNet-53                       & 77.63                     & 80.45                     & 72.30                      & 63.59                     & 68.36                     & 56.81                      & 67.30                     & 67.20                      \\ 
\hline\hline
\textit{Transformer-based}                       & \multicolumn{1}{l|}{}  & \multicolumn{1}{l|}{}            & \multicolumn{1}{l}{}      & \multicolumn{1}{l}{}      & \multicolumn{1}{l|}{}      & \multicolumn{1}{l}{}      & \multicolumn{1}{l}{}      & \multicolumn{1}{l|}{}      & \multicolumn{1}{l}{}      & \multicolumn{1}{l}{}       \\
\textit{Pretrained:}                             & \multicolumn{1}{l|}{}  & \multicolumn{1}{l|}{}            & \multicolumn{1}{l}{}      & \multicolumn{1}{l}{}      & \multicolumn{1}{l|}{}      & \multicolumn{1}{l}{}      & \multicolumn{1}{l}{}      & \multicolumn{1}{l|}{}      & \multicolumn{1}{l}{}      & \multicolumn{1}{l}{}       \\
ViLBERT~\cite{ViLBERT}                                         & NeurIPS'19                & ResNet-101                       & -                         & -                         & -                          & 72.34                     & 78.52                     & 62.61                      & -                         & -                          \\
ERNIE-ViL~\cite{ERNIE-ViL}                                      & AAAI'20                & ResNet-101                       & -                         & -                         & -                          & 75.95                     & 82.07                     & 66.88                      & -                         & -                          \\
UNTIER~\cite{UNITER}                                         & ECCV'20                & ResNet-101                       & 81.41                     & 87.04                     & 74.17                      & 75.90                     & 81.45                     & 66.70                      & 74.86                     & 75.77                      \\
VILLA~\cite{VILLA}                                          & NeurIPS'20                & ResNet-101                       & 82.39                     & 87.48                     & 74.84                      & 76.17                     & 81.54                     & 66.84                      & 76.18                     & 76.71                      \\
MDETR~\cite{MDETR}                                           & ICCV'21                & ResNet-101                       & \multicolumn{1}{l}{86.75}           & \multicolumn{1}{l}{89.58} & \multicolumn{1}{l|}{81.41}           & \multicolumn{1}{l}{79.52} & \multicolumn{1}{l}{84.09} & \multicolumn{1}{l|}{70.62} & \multicolumn{1}{l}{81.64} & \multicolumn{1}{l}{80.89} \\
\hline\hline
\multicolumn{1}{l|}{\textit{~~~Transformer-based} } & \multicolumn{1}{l|}{}  & \multicolumn{1}{l|}{}            & \multicolumn{1}{l}{}      & \multicolumn{1}{l}{}      & \multicolumn{1}{l|}{}      & \multicolumn{1}{l}{}      & \multicolumn{1}{l}{}      & \multicolumn{1}{l|}{}      & \multicolumn{1}{l}{}      & \multicolumn{1}{l}{}       \\
\textit{without Pretrained:}                     & \multicolumn{1}{l|}{}  & \multicolumn{1}{l|}{}            & \multicolumn{1}{l}{}      & \multicolumn{1}{l}{}      & \multicolumn{1}{l|}{}      & \multicolumn{1}{l}{}      & \multicolumn{1}{l}{}      & \multicolumn{1}{l|}{}      & \multicolumn{1}{l}{}      & \multicolumn{1}{l}{}       \\
TransVG~\cite{TransVG}                                         & ICCV'21                & ResNet-101                       & 81.02                     & 82.72                     & 78.35                      & 64.82                     & 70.70                     & 56.94                      & 68.67                     & 67.73                      \\
MDETR (w/o pretrain)~                                               & ICCV'21                & ResNet-101                       & 78.01                     & 82.18                     & 72.56                                & 68.01                     & 72.83                     & 55.57                      & 65.54                     & 65.99                      \\
MDETR*                                                              & -                      & ResNet-101                       & \multicolumn{1}{l}{81.49} & \multicolumn{1}{l}{84.67} & \multicolumn{1}{l|}{76.58}           & \multicolumn{1}{l}{70.93} & \multicolumn{1}{l}{75.65} & \multicolumn{1}{l|}{59.27} & \multicolumn{1}{l}{69.59} & \multicolumn{1}{l}{70.22}  \\
\textbf{MDETR* + SiRi}                                                & -                      & ResNet-101                       & \multicolumn{1}{l}{\textbf{85.83}} & \multicolumn{1}{l}{88.56} & \multicolumn{1}{l|}{\textbf{81.27} } & \multicolumn{1}{l}{76.68} & \multicolumn{1}{l}{82.01} & \multicolumn{1}{l|}{66.33} & \multicolumn{1}{l}{76.63} & \multicolumn{1}{l}{76.46}  \\
\textbf{MDETR* + MT SiRi}                                             & -                      & ResNet-101                       & 85.82            & \textbf{89.11}            & 81.08                                & \textbf{77.47}            & \textbf{83.04}            & \textbf{67.11}             & \textbf{77.39}            & \textbf{76.80}             \\
\bottomrule
\end{tabular}
}

\label{tab:sota_compare}
\end{table}

\subsection{Experimental Settings}
\label{sec:expset}

\textbf{Implementation Details.}
Following MDETR~\cite{MDETR}, all parameters in the network are optimized using AdamW~\cite{AdamW} with the learning rate warm-up strategy.
The model is trained using 4 GPUs with a batch size of 72.
We set the learning rate of the language backbone RoBERTa~\cite{RoBERTa} to be $1\times10^{-5}$, and all the rest parameters to be $5\times10^{-5}$.
In initial training, the model with a single decoder is trained for 55 epochs, and the model with a dual decoder (multi-task SiRi) is trained for 35 epochs since it converges quickly. Each retraining stage takes another 30 training epochs. 
We set the maximum side length of the input image as 640 while keeping the original aspect ratio. Images in the same batch are padded with zeros until acquiring the largest size of that batch. Similarly, sentences in one batch will be adjusted to the same length as well. 
We continually retrain the model until the validation performance converges (usually 5 to 8 rounds).

\textbf{Evaluation Metrics. }
Following the proposal setting in the previous work, we use the metric Prec@0.5 to evaluate our method, where a predicted region will be regarded as a positive sample if its intersection over union (IoU) with the ground-truth bounding box is greater than 0.5.

\subsection{Comparison with State-of-the-art Methods}
\label{sec:sota}

We compare our method with other state-of-the-art methods on three common benchmarks of Referring Expression Comprehension, \ie, RefCOCO, RefCOCO+, and RefCOCOg.
Results are reported in Table~\ref{tab:sota_compare}.  Our method displays significant improvement over previous methods on all three datasets. 
Compared to models without large-scale pretraining, which is a fair comparison, we outperform them by more than 6.39\% on RefCOCO@testA, 10.21\% on RefCOCO+@testA, and 9.07\% on  RefCOCOg@test.
Even compared to those large-scaled pretrained models, \eg, MDETR pretrained using more than one million aligned image-text pairs, our method still achieves comparable results on RefCOCO without those extra data.

\subsection{Ablation Studies}
\label{sec:ablation}

\textbf{Different Retraining Module.}
Besides continually updating the encoder while periodically re-initializing all the other parts, we also evaluate different re-initializing modules.
\begin{figure*}[!t]
  \begin{minipage}[b]{0.70\textwidth} 
        \begin{center}
        \includegraphics[width=0.99\linewidth]{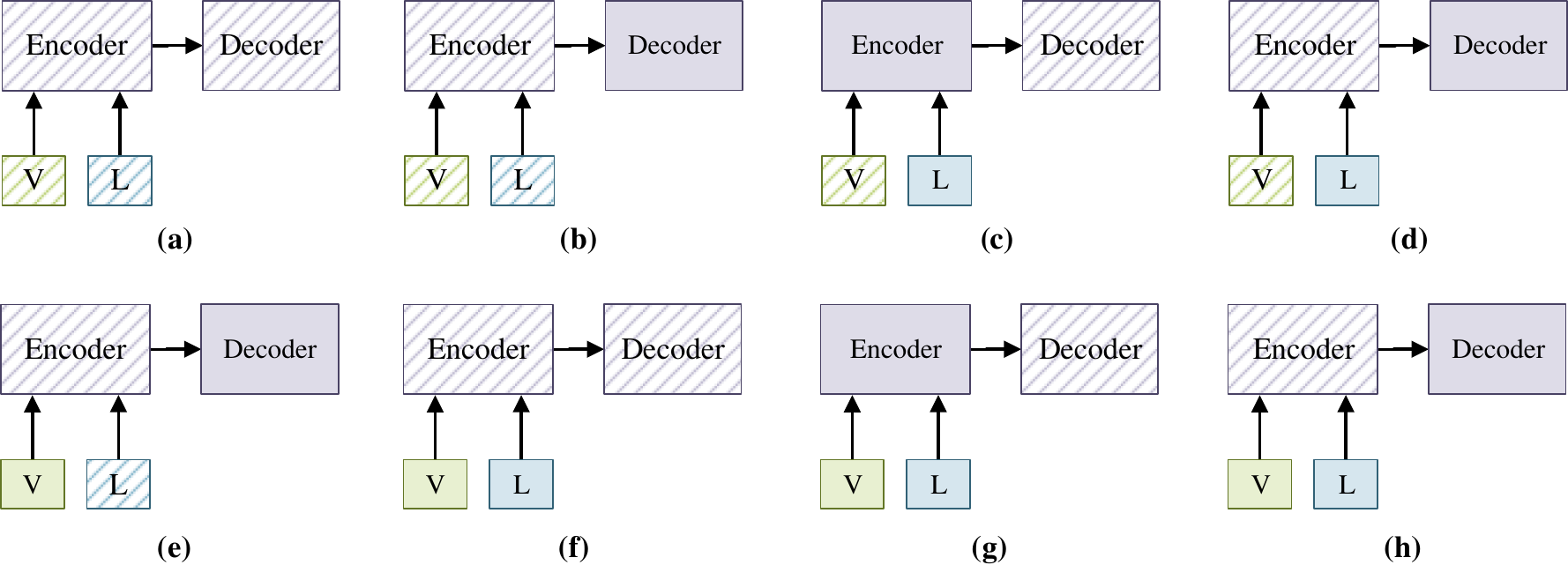}
        \end{center}
          \caption{Schematic of the eight retraining variants with different combinations of selective modules. The solid color background means re-initializing parameters, while the slash background means continually updated parameters from previous periods. Best viewed in color. }
        \label{fig:retrain mode}
  \end{minipage}%
  \hspace{0.01\textwidth} 
  \begin{minipage}[b]{0.28\textwidth} 
    \centering
    \small
    \scalebox{0.75}{
    \begin{tabular}{l|c} 
    \toprule
    \multicolumn{1}{c|}{Mode} & \multicolumn{1}{c}{RefCOCO+@val}  \\
    \hline\hline
    \multicolumn{1}{c|}{Initial}              & 71.45        \\ 
    \hline 
    (a) V,L,E,D                                & 71.98         \\
    (b) V,L,E                                  & 72.12         \\
    (c) V,D                                    & 72.14         \\
    (d) V,E                                    & 73.25          \\
    (e) L,E                                    & 73.44          \\
    (f) E,D                                    & 73.80          \\
    (g) D                                      & 72.76          \\
    (h) E                                      & \textbf{74.14}         \\
    \bottomrule
    \end{tabular}}
    \makeatletter\def\@captype{table}\makeatother 
        \setlength{\abovecaptionskip}{5mm} 
        \caption{Performance comparison of different selective modules. The eight mode are shown in Fig.~\ref{fig:retrain mode}.}
  
    \label{tab:retrain}
  \end{minipage} 
\end{figure*}

We show eight variants of our SiRi Mechanism in Fig.~\ref{fig:retrain mode}, 
For a fair comparison, we keep all hyperparameters the same and retrain these variants from the same initial trained model.
We show their correspondence results after the first retraining in Table~\ref{tab:retrain}.
The encoder with better initialized parameters is the critical factor for the  whole model converging to a better local minimum.

Comparing mode (d) with mode (h), we find that re-initializing the \textit{visual} backbone has great impact on performance boosting, which verifies our motivation that re-initializing the input of encoder helps to get out of local minimums while keeping the essential cross-modeling ability of previous models.
Similar results can be found for \textit{language} backbone by comparing mode (e) with mode (h).
Interestingly, we find that the performance is competitive to Mode (h) when we use Mode (f), where we keep the parameters of both encoder and decoder.
For simplicity, we only keep the encoder updated continually in all the other experiments.

\begin{figure*}[!t]
  \begin{minipage}[b]{0.66\textwidth} 
        \begin{center}
        \includegraphics[width=0.99\linewidth]{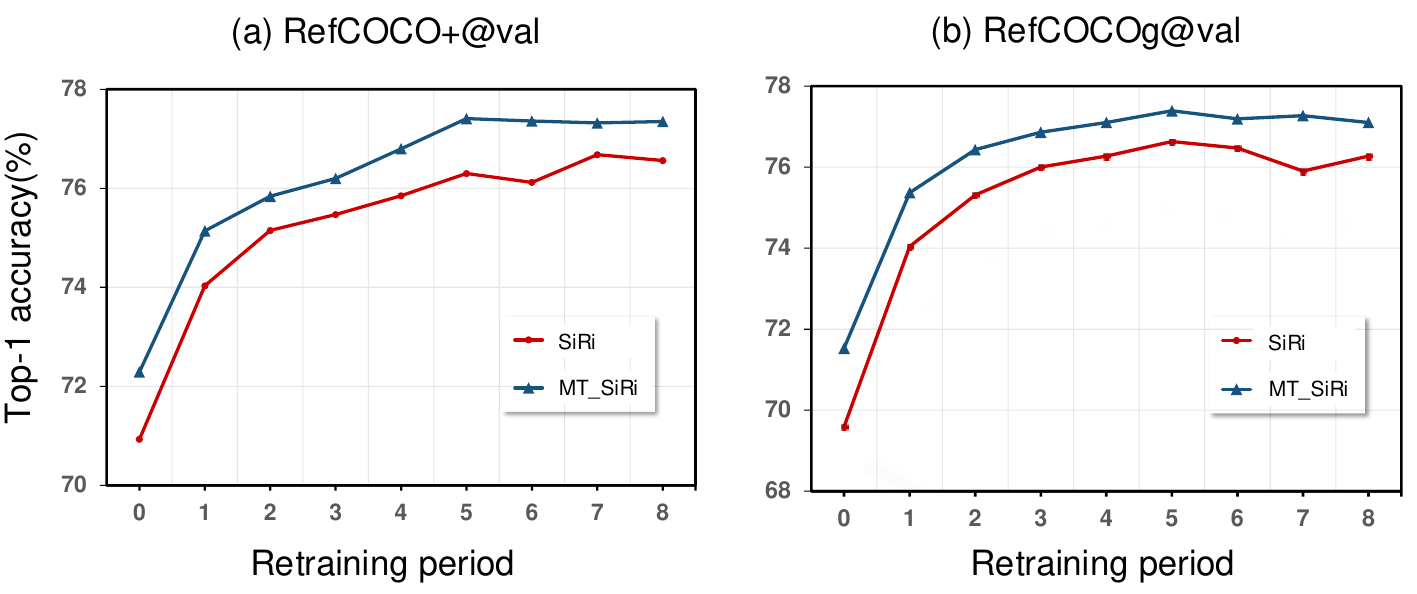}
        \end{center}
          \caption{Performance achieved by increasing the training periods. The blue line indicates the single SiRi model and the red line indicates the multi-task SiRi model. ``MT" indicates multi-task.}
          \label{fig:retrain_times}
  \end{minipage}%
  \hspace{0.01\textwidth} 
  \begin{minipage}[b]{0.32\textwidth} 
    \centering
    \small
    \scalebox{0.67}{
    \begin{tabular}{c|cc|c} 
    \toprule
    \multirow{2}{*}{Structure}      & \multicolumn{2}{c|}{Object Queries} & \multirow{2}{*}{RefCOCO+}  \\
                                    & \textit{1st} Dec.  & \textit{2nd} Dec.                 &                                \\ 
    \hline\hline
    \multirow{2}{*}{Single-task} & L & –                        & 70.93                          \\
                                    & C  & –                        & 70.72                          \\ 
    \hline
    \multirow{3}{*}{Multi-task}   & L & L                & 70.27                          \\
                                    & C  & C                 & 71.24                          \\
                                    & L & C                 & \textbf{72.29}                          \\
    \bottomrule
    \end{tabular}}
    \makeatletter\def\@captype{table}\makeatother %
        \setlength{\abovecaptionskip}{6mm} 
        \caption{Ablation studies on different object query types in multi-task SiRi. (``L": learnable queries, and ``C": constant queries, ``Dec.": Decoder.)}
    \label{tab:query_set}
  \end{minipage} 
\end{figure*}

\textbf{Retraining Periods.}
In Fig.~\ref{fig:retrain_times}, we show the validation performance curves during selective retraining.
Zero indicates the initial trained model in the figure.
We can see the model performance increases a lot in the first three retraining periods and then tends to converge after several retraining periods.
The highest performances are achieved in the fifth retraining period, where SiRi outperforms the initial trained model by 5.18\% (72.29\% versus 77.47\%) and 5.86\% (71.53\% versus 77.39\%) on RefCOCO+ and RefCOCOg, respectively.

\textbf{Different Object Queries in Multi-task SiRi.} We can also see the consistent performance gap between the single SiRi and the multi-task SiRi in Fig.~\ref{fig:retrain_times}.
The multi-task SiRi always performs better than single SiRi during all the retraining periods.
We further study the impact of different object queries (\eg, learnable queries and constant queries) used in Multi-task SiRi.
The results of the initial trained models using different quires in multi-task learning are shown in Table~\ref{tab:query_set}. 

Although learnable and constant object queries achieve similar results for single task training, the combination of them in multi-task learning achieves higher performance (72.29\% \textit{versus} 70.93\% on RefCOCO+).
Note that multi-task structure with two identical object query types (\eg, both learnable or both constant) does not outperform single task learning.
It indicates that taking different queries to attend the same encoder output may help the encoder to be more robust on vision-language interaction.

\subsection{Qualitative Results}
\label{sec:visualization}

\begin{figure*}[t]
\begin{center}
\includegraphics[width=0.95\linewidth]{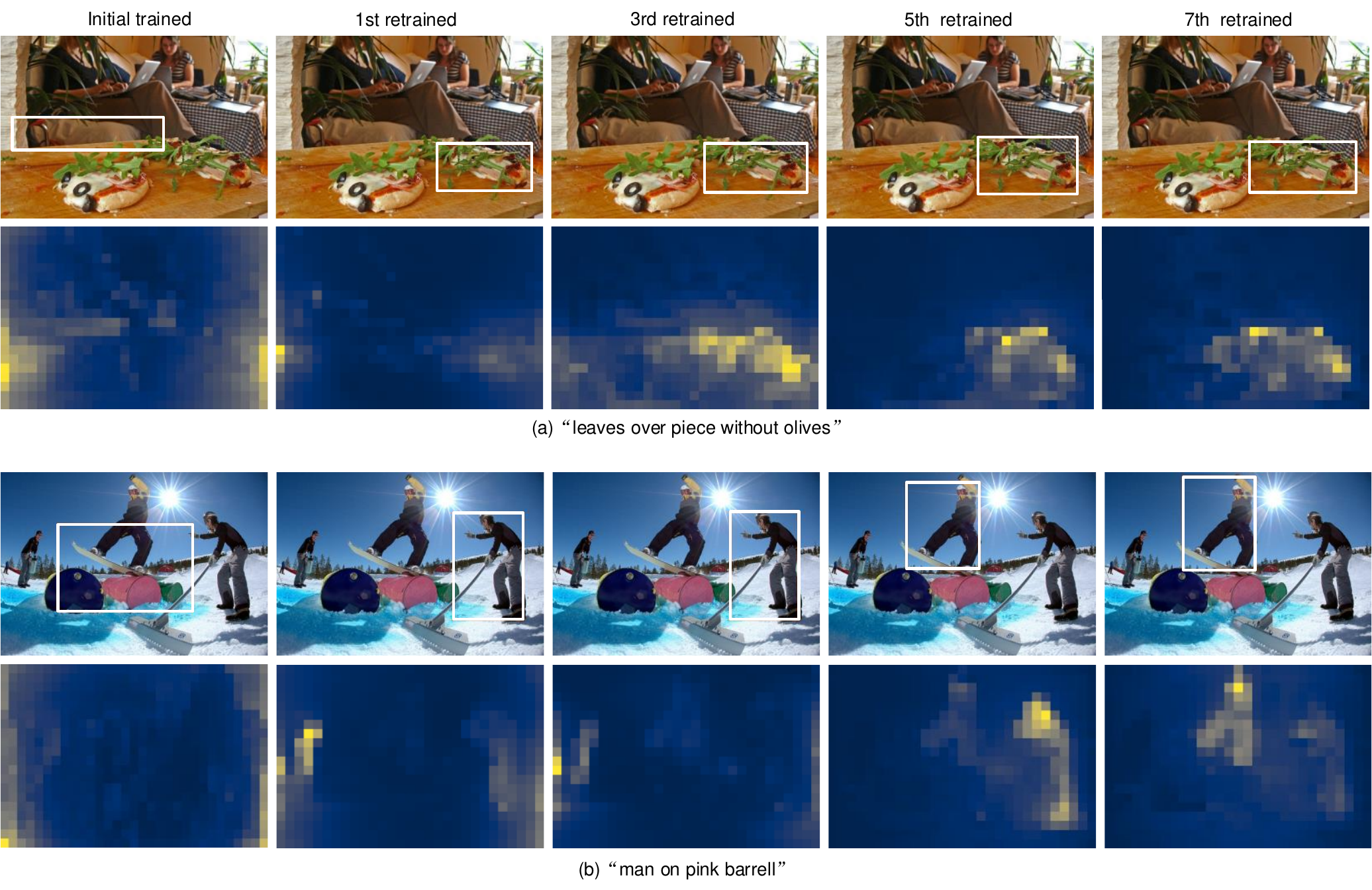}
\end{center}
   \caption{Visualization of the predicted box and the encoder's cross-modal attention weights in inference. The columns represent initial trained, 1st retrained, 3rd retrained, 5th retrained, 7th retrained model, respectively, from left to right. As we can see, the model prediction gets better as the encoder attention map gets clear.
   }
\label{fig:attention visual}
\end{figure*}

We visualize the attention weight of encoders along with the retraining progress in Fig.~\ref{fig:attention visual}.
To be specific, we calculate the cross-modal attention weights (vision output tokens based on language input tokens) from the last layer of the Transformer encoder, and then visualize them in the original image size.
We believe the values of cross-modal attention weights indicate the encoder's ability of vision-language understanding. 

We show two test samples in the figure with the corresponding input sentences.
From left to right, we show the bounding box predictions together with the attention maps generated by the initial trained, 1st, 3rd, 5th, and 7th retrained encoders, respectively.
It can be intuitively seen that the encode learns to better perceive the relationship between expressions and images as the continuous SiRi training goes.
Taking the upper sample as an example, the predicted bounding box is incorrect from the initial trained model, where we can see the attention map of the first encoder does not highlight the referred object, either.

After selective retraining, the encoder gets better and better, which can be seen from the more accurate attention maps.
Therefore, the predicted boxes are also better than the initial ones.
It validates our motivation that the better encoder initialization helps the model converge to a better local minimum. 
Continually updating the encoder while periodically re-initializing other modules can strengthen the visual-linguistic modeling.

\subsection{Extensibility Studies}
\label{sec:extensibility}

To better show the generality, we further extend SiRi to more visual grounding settings, models, and tasks.

\begin{wrapfigure}[17]{r}{6cm}
\centering
\includegraphics[width=0.95\linewidth]{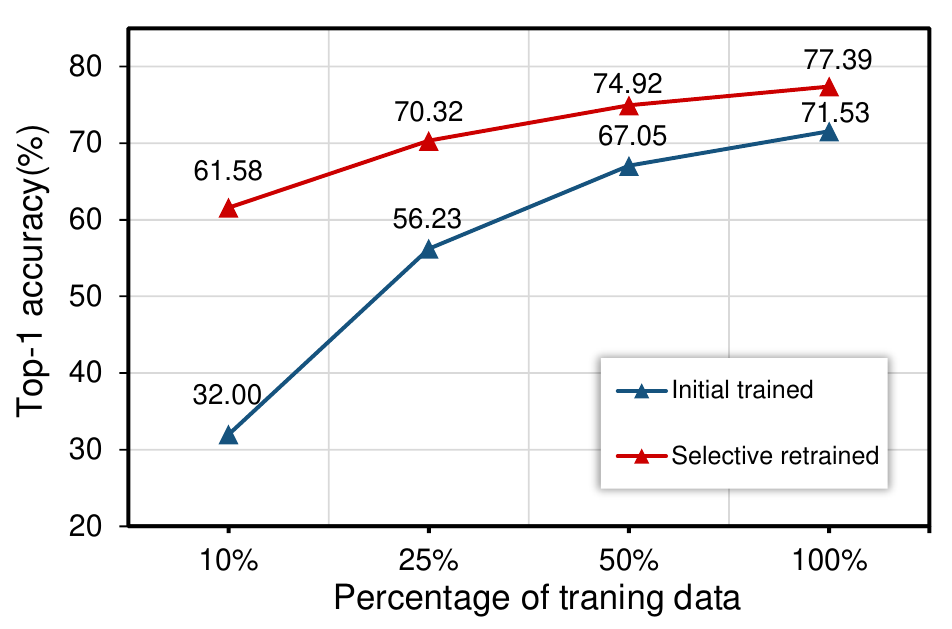}
\caption{Performance improvement of the model with SiRi with limited training samples. We randomly sample 10\%, 25\%, 50\% of training data from RefCOCOg and train with SiRi. All models are evaluated on the same RefCOCOg \textit{val} set. }
\label{fig:datasets_scale}
\end{wrapfigure}

\textbf{Extend to Small Data Size.}
First, we study how SiRi performs with fewer training data, where the over-fitting issue is more severe. To do so, we randomly sample 10\%, 25\%, and 50\% of training data from the RefCOCOg training set as the new training splits, respectively.
Then we train the model following the SiRi mechanism\footnote{We train more epochs until converging in small-scale experiments.} and then evaluate the performance on the full validation set of RefCOCOg (the same validation set for all).
The results are shown in Fig.~\ref{fig:datasets_scale}.
Compared with the initial trained model, our SiRi model shows very impressive performance gains, \eg, almost doubling the performance at 10\% sampling rate.

As can be seen from the figure, the performance is improved much more significantly when employing the SiRi mechanism on fewer training data, which verifies that our SiRi can generalize the vision-language encoder and avoid over-fitting.
It suggests that our SiRi mechanism may be potentially treated as a strong alternative to large-scale pre-training models.


\noindent \textbf{Extend to other V-L models.}
{The application of SiRi mechanism on other V-L models can be achieved by simply following the principle: keeping the parameters of V-L fusion module continuously training, while reinitializing the other parts. We applied our SiRi to Transformer-based Visual Grounding model TransVG~\cite{TransVG} and RES model LAVT~\cite{yang2022lavt}. Experimental details are presented in 
Appendix~\ref{subsec:transvg_lavt}.}
For TransVG~\cite{TransVG}, we report REC and Phrase Grounding results in Table~\ref{tab:TransVG}. We found that SiRi could further improve the performance of TransVG by an average of 2\% at top-1 accuracy on all four REC datasets, and the performance has also been effectively improved on Phrase Grounding dataset Flickr30k dataset. For LAVT~\cite{yang2022lavt}, We report the results of SiRi in RES dataset RefCOCO+ three splits \textit{val}, \textit{testA}, \textit{testB} in Table~\ref{tab:lavt_result}.

\begin{table}[!t]
\centering
\caption{REC and phrase grounding results of TransVG~\cite{TransVG} with SiRi mechanism.}
\scalebox{0.85}{
\begin{tabular}{c|c|ccc|ccc|c|cc|cc} 
\toprule
\multirow{3}{*}{Model} & \multirow{3}{*}{Backbone} & \multicolumn{9}{c|}{Referring Expression Comprehension}                                                                                                       & \multicolumn{2}{c}{PhraseGround}  \\ 
\cline{3-13}
                       &                           & \multicolumn{3}{c|}{RefCOCO}                        & \multicolumn{3}{c|}{RefCOCO+}                       & RefCOCOg        & \multicolumn{2}{c|}{ReferIt}    & \multicolumn{2}{c}{Flickr30k}        \\
                       &                           & val             & testA           & testB           & val             & testA           & testB           & g-val           & val            & test           & val            & test                \\ 
\hline\hline
TransVG                & ResNet-50                 & 80.49           & 83.28           & 75.24           & 66.39           & 70.55           & 57.66           & 66.35           & 71.60          & 69.76          & 77.19          & 78.47               \\
+SiRi                  & ResNet-50                 & \textbf{82.97}  & \textbf{84.42}  & \textbf{79.04}  & \textbf{69.30}  & \textbf{73.27}  & \textbf{59.93}  & \textbf{68.54}  & \textbf{74.28} & \textbf{71.36} & \textbf{77.99} & \textbf{79.17}      \\
\bottomrule
\end{tabular}}
\label{tab:TransVG}
\end{table}

\begin{table}[!t]
\centering
\caption{Referring Expression Segmentation results of LAVT~\cite{yang2022lavt} with SiRi.}
\scalebox{0.8}{
\setlength{\tabcolsep}{3.5mm}{
\begin{tabular}{c|c|ccccc|c|c} 
\toprule
RefCOCO+               & Model & P@0.5          & P@0.6          & P@0.7          & P@0.8          & P@0.9          & oIoU           & mIoU            \\ 
\hline\hline
\multirow{2}{*}{val}   & LAVT  & 74.44          & 70.91          & 65.58          & 56.34          & 30.23          & 62.14          & 65.81           \\
                      & +SiRi & \textbf{75.56} & \textbf{72.39} & \textbf{67.88} & \textbf{58.33} & \textbf{30.79} & \textbf{62.86} & \textbf{66.78}  \\ 
\hline
\multirow{2}{*}{testA} & LAVT  & 80.68          & 77.96          & 72.90          & 62.21          & 32.36          & 68.38          & 70.97           \\
                      & +SiRi & \textbf{82.20} & \textbf{79.18} & \textbf{74.54} & \textbf{63.99} & \textbf{32.62} & \textbf{68.87} & \textbf{71.93}  \\ 
\hline
\multirow{2}{*}{testB} & LAVT  & 65.66          & 61.85          & 55.94          & 47.56          & 27.24          & \textbf{55.10} & 59.23           \\
                      & +SiRi & \textbf{66.41} & \textbf{62.86} & \textbf{57.37} & \textbf{49.23} & \textbf{27.90} & 55.03          & \textbf{59.70}  \\
\bottomrule
\end{tabular}}}
\label{tab:lavt_result}
\end{table}


\noindent \textbf{Extend to other V-L tasks.} We also test our SiRi in more vision-language tasks, including referring expression segmentation, phrase grounding, and visual question answering.
For these experiments, we took the transformer-based MDETR model (without pre-training) as our baseline. 
The specific settings of how to apply SiRi on these tasks are stated as follows.

\noindent \textbf{\textit{-Referring Expression Segmentation (RES).}} RES is to segment the objects according to the given language description. We further perform the segmentation task on the trained visual grounding model. We keep the original MDETR model architecture the same but modify the hyperparameters according to the settings used in training visual grounding in this paper. 
We test the SiRi model on three RES datasets, \ie, RefCOCO, RefCOCO+, RefCOCOg.
In Table \ref{tab:res}, we report the RES performance of the SiRi model after \textit{Initial-train}, \textit{3rd-train}, and \textit{5th-train} stages.
It can be seen that SiRi can steadily improve RES models during the retraining process.

\begin{table}
\centering
\caption{Experiment results on RES. We report precision Pr@0.5, 0.7, 0.9 and overall IoU on the \textit{val} set of RefCOCO, RefCOCO+, RefCOCO.}
\label{tab:res}
\scalebox{0.8}{
\begin{tabular}{c|cccc|cccc|cccc} 
\toprule
\multirow{2}{*}{Stage} & \multicolumn{4}{c|}{RefCOCO}     & \multicolumn{4}{c|}{RefCOCO+}    & \multicolumn{4}{c}{RefCOCOg}      \\
                       & Pr@0.5 & Pr@0.7 & Pr@0.9 & oIoU  & Pr@0.5 & Pr@0.7 & Pr@0.9 & oIoU  & Pr@0.5 & Pr@0.7 & Pr@0.9 & oIoU   \\ 
\hline\hline
Initial-train          & 77.76  & 68.89  & 28.58  & 62.12 & 68.36  & 61.11  & 25.89  & 52.48 & 64.34  & 54.84  & 20.42  & 51.39  \\
3rd-retrain            & 82.58  & 74.33  & 32.57  & 68.02 & 75.27  & 67.76  & 28.21  & 60.11 &  72.20      &61.46        &25.12        &58.33        \\
5th-retrain            & \textbf{83.56}  & \textbf{75.37}  & \textbf{32.79}  & \textbf{69.34} & \textbf{76.46}  & \textbf{68.47}  & \textbf{28.26}  & \textbf{61.1}5 & \textbf{73.24} & \textbf{63.25}   &  \textbf{25.08}      & \textbf{59.69}       \\
\bottomrule
\end{tabular}}
\end{table}

\noindent \textbf{\textit{-Phrase Grounding.}}
The task is to locate objects in an image based on the phrases which may be inter-related.
We evaluate the SiRi mechanism on the Flickr30k entities dataset. For the input image, we set the maximum size to 800.  We show the model performance of different SiRi stages in Table \ref{tab:pg_gqa}.
We can see SiRi further improves the initial trained model by 1\%$\sim$2\% on Recall@1, Recall@5, Recall@10 (denoted as R@1, R@5, R@10, respectively).

\noindent \textbf{\textit{-Visual Question Answering.}} Given an image and a question in natural language, this task is to infer the correct answer. We use the scene graph provided in GQA to align question words and the boxes as in MDETR. We verify the validity of SiRi on the visual question answering task in GQA \textit{balanced} split dataset. 
The results of SiRi model from different training stages are reported in Table \ref{tab:pg_gqa}. The accuracy is improved from 55.75 to 57.45. 

\begin{table}[!t]
\centering
\small
\caption{Experiment results of Phrase Grounding on the validation set of Flickr30k and the VQA performance on the GQA \textit{balance test} set.}
\scalebox{0.85}{
\setlength{\tabcolsep}{8.5mm}{
\begin{tabular}{c|ccc|c} 
\toprule
\multirow{2}{*}{Stage} & \multicolumn{3}{c|}{Phrase Grounding@Flickr30k} & GQA  \\
                       & R@1   & R@5   & R@10                            & Accuracy     \\ 
\hline\hline
Initial-train          & 76.22 & 87.19 & 90.26                           & 55.75    \\
1st-retrain            & 78.41 & 88.42 & 91.31                           & 56.38    \\
2nd-retrain            & \textbf{78.63} & \textbf{88.62} & \textbf{91.62}                           & \textbf{57.25}    \\
\bottomrule
\end{tabular}}}
\label{tab:pg_gqa}
\end{table}

\section{Conclusion}
In this paper, we present a novel training mechanism namely Selective Retraining (SiRi) for visual grounding, where we keep updating the Transformer encoder while re-initialize the other modules to get out of local minimums.
We further propose multi-task SiRi to train a better encoder by incorporating an auxiliary decoder with constant input queries. 
Extensive experiments prove our method helps the Transformer encoder better perceive the relationship between the visual and the corresponding expression, outperforming state-of-the-art methods on the three visual grounding datasets.
Interestingly, we find SiRi also performs superior even with very limited training data. 
Even with a quarter of training data, we outperform state-of-the-art methods (with full training data) by 1.65\% on the RefCOCOg validation set.
We also extend SiRi to other Transformer-based visual grounding models and other V-L tasks. We hope our work will help motivate more researchers in the V-L research community in the future.

\textbf{Acknowledgements.}
This work was supported in part by the National Key R\&D Program of China (No.2021ZD0112100), the National NSF of China (No.U1936212, No.62120106009), the Fundamental Research Funds for the Central Universities (No.K22RC00010). We thank Princeton Visual AI Lab members (Dora Zhao, Jihoon Chung, and others) for their helpful suggestions.

\newpage
{\small
\bibliographystyle{ieee_fullname}
\bibliography{egbib}
}

\newpage
\appendix
\renewcommand{\appendixname}{Appendix~\Alph{section}}

\section{Details of Multi-task SiRi}
\label{subsec:dual decoder details}

\begin{figure}
\begin{center}
\includegraphics[width=0.97\linewidth]{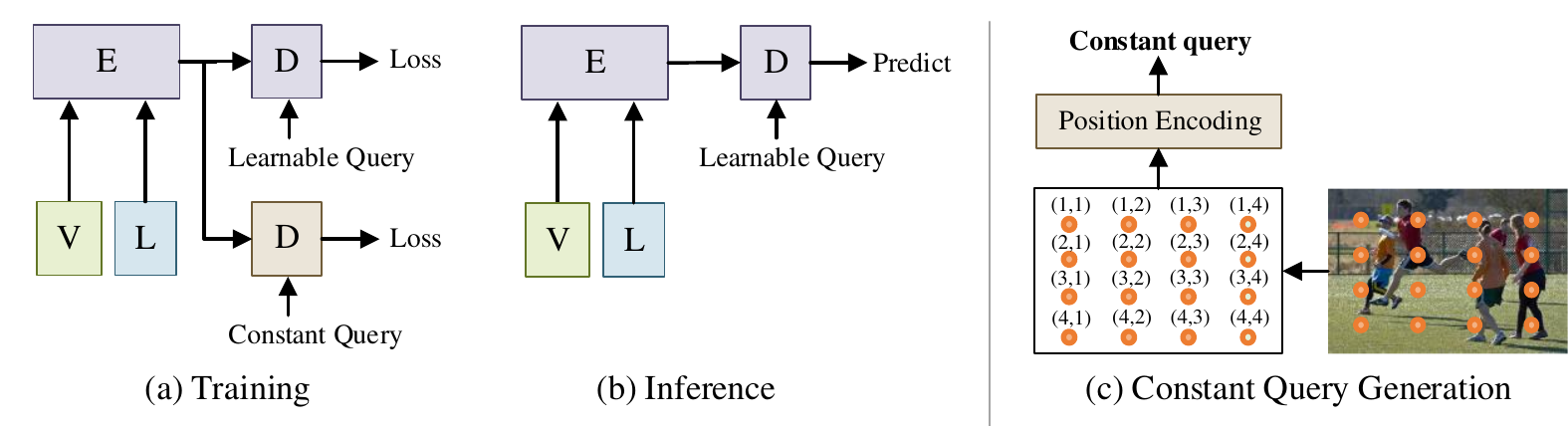}
\end{center}
  \caption{Overview of the multi-task SiRi. 
  We generate constant query as shown in (c). }
\label{fig:decoder}
\end{figure}

As shown in Fig.~\ref{fig:decoder}, in multi-task SiRi, we leverage an auxiliary decoder (no weights sharing) for multi-task learning in each training/retraining stage. 
The losses of two decoders are summed up as the overall objective function for optimization.

After training/retraining, the auxiliary decoder was dropped after training so that we keep the same amount of parameters and operations (inference speed) in model inference.

In detail, we generate constant grid points by dividing the image into patches. Then we take the grid intersections for position encoding, as shown in Fig.~\ref{fig:decoder}.
The coordinates of the $k$-th intersection point $P_{k}$ are,
\begin{equation}
P_{k}= (\frac{k_1}{\sqrt{n}+1 } , \frac{k_2}{\sqrt{n}+1 }), 
k_1, k_2 \in \{ 1, 2, ..., \sqrt{n}  \},
\end{equation}
where $n$ is the number of object queries. Based on the generated constant points $P$, the constant queries ${Q}_{c}$ can be formulated as follows,
\begin{equation}
Q_{c}=
\begin{cases}
 {PE(P,2i)= sin(\frac{P}{10000^{2i/C}})}
 \\
 {PE(P,2i+1)= cos(\frac{P}{10000^{2i/C}})},
\end{cases}
\end{equation}
where $C$ denotes the dimension of the query embedding, and $i$ is the dimension index.

Therefore, in multi-task SiRi, we leverage an auxiliary decoder (no weights sharing) for multi-task learning in each training/retraining stage. This auxiliary decoder was dropped after training so that we keep the same amount of parameters and operations in model inference.

During training, the model weights of the two decoders are randomly initialized and separately updated. In other words, they do not share weights. 
We individually calculate the loss on each decoder's prediction and then simply add the two losses as the overall objective function for optimization.
For inference, we can keep \textit{either one} of the two trained decoders and take its prediction as the final prediction. 

Thus, the inference speed is exactly the same as the previous single decoder framework.

We found in experiments that both decoders in the multi-task structure achieve very similar performance and significantly outperform either of them in the previous single-task framework. 
This proves that the performance gains are from better-optimized encoders, rather than additional computation or model parameters.

\section{Additional Experimental Analysis}
\subsection{More Experiment Results and Setting Details of SiRi in other V-L models}
\label{subsec:transvg_lavt}
\begin{figure*}
  \centering
    \includegraphics[width=0.99\linewidth]{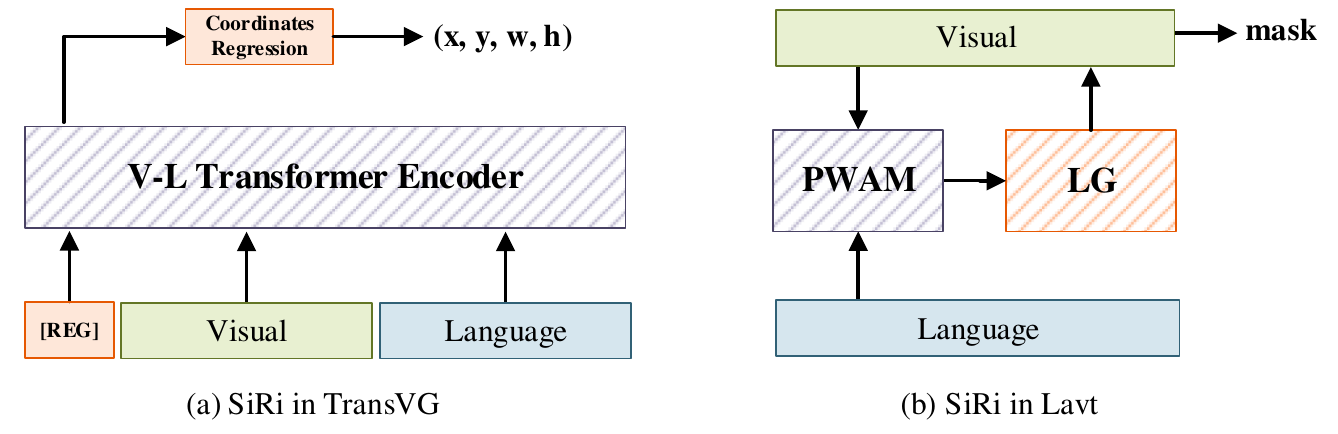}
   \caption{The illustration of (a) TransVG~\cite{TransVG}  and (b) LAVT~\cite{yang2022lavt} with SiRi. The solid color background means re-initializing, while the slash color background means continually updating.
   }
   \label{fig:illustration}
\end{figure*}

As shown in Fig.~\ref{fig:illustration} (a), we apply SiRi mechanism by keeping the V-L Transformer encoder continually trained while the other modules re-initialized. The [REG] token is used to enter the coordinates regression module and locate the object. We strictly adopted the same hyper-parameters as TransVG. 
In LAVT~\cite{yang2022lavt}, Pixel-Word Attention Module (PWAM) and Language Gate (LG) are modules for V-L interaction. As shown in Fig.~\ref{fig:illustration} (b), we keep the two module parameters continuously trained and reinitialize the other parameters. 

\vspace{-5mm}

\vspace{-4mm}

\subsection{Additional Qualitative Results}
\begin{figure*}[t]
  \centering
    \includegraphics[width=1\linewidth]{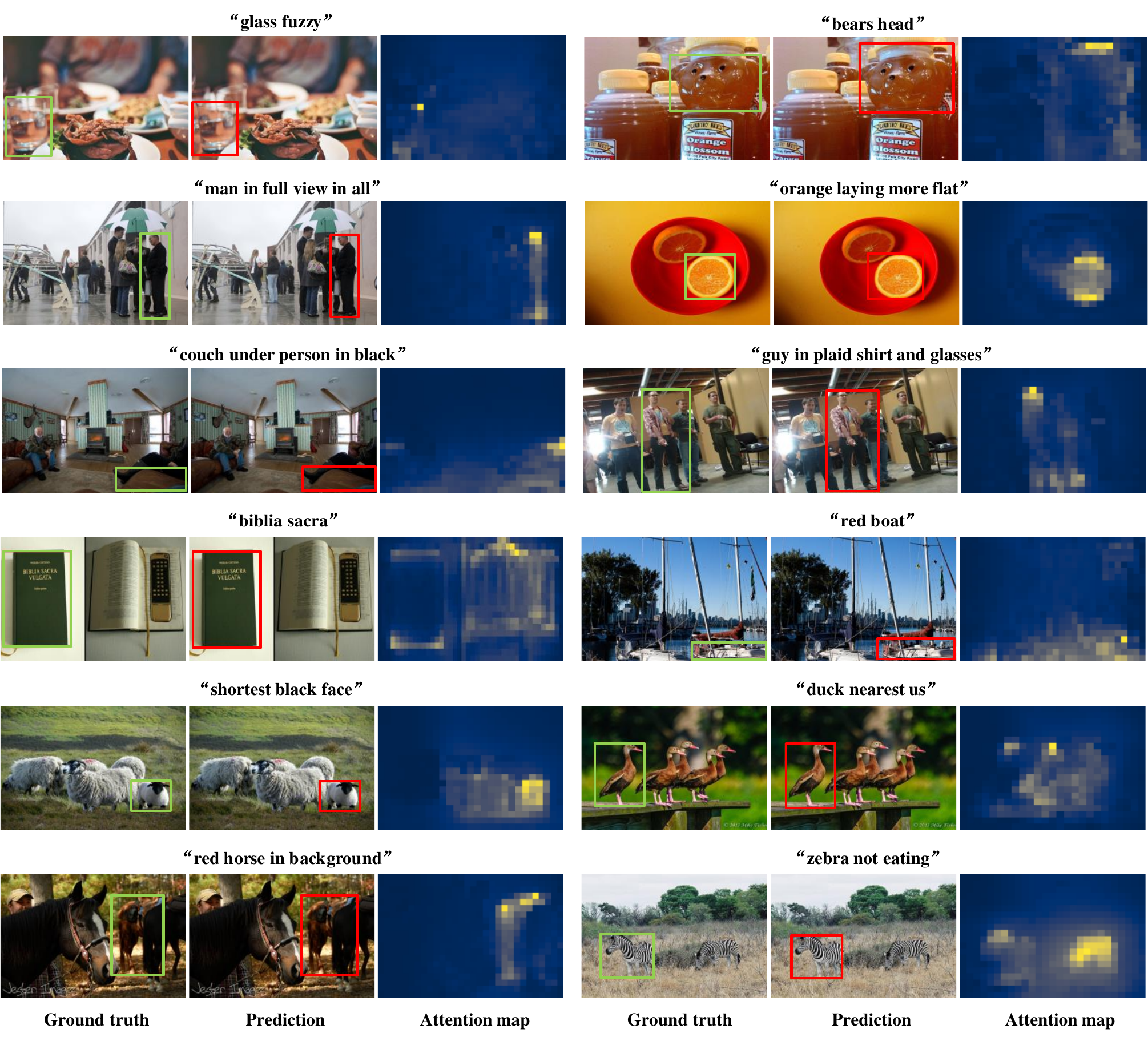}
   \caption{Examples of correct comprehension of referring expressions on RefCOCO+.
   }
   \label{fig:examples}
\end{figure*}

\begin{figure*}
  \centering
    \includegraphics[width=1\linewidth]{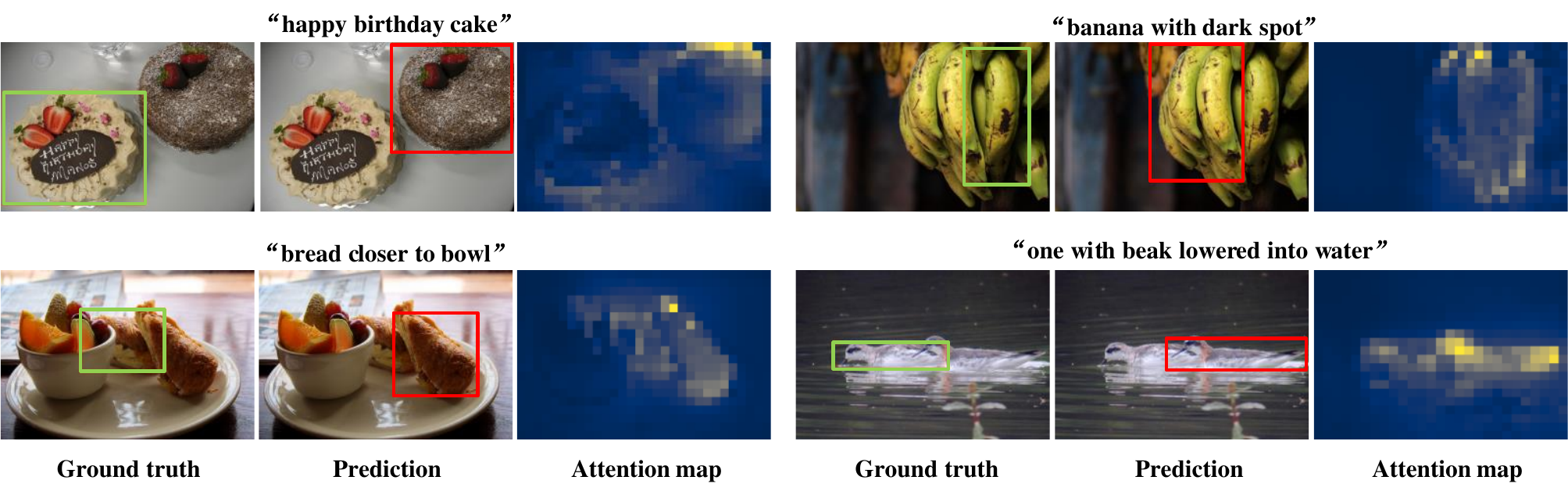}
   \caption{Failure cases of our model prediction on RefCOCO+.
   }
   \label{fig:false examples}
\end{figure*}

We show more qualitative results of our trained model in Fig.~\ref{fig:examples} and Fig.~\ref{fig:false examples}. Each example set includes the ground truth (the left one), prediction of our method (the middle one), and the attention map of the encoder (the right one). The green box indicates the ground truth annotation, while the red one represents the prediction box of our trained model. Fig.~\ref{fig:examples} shows some correct prediction examples of referring expression comprehension, while Fig.~\ref{fig:false examples} contains several incorrect predictions. These visualization examples demonstrate that our approach can model the relative relationship description, \eg, the relationship of \textbf{``couch"} and \textbf{``person"} in ``couch under person in black". 
In addition, we can also find that the attention map of the encoder tends to be more attentive to the object referred to by the expression (with higher brightness).

\subsection{Training Loss}
\begin{wrapfigure}[20]{r}{6cm}%
\centering
\vspace{-3em}
\includegraphics[width=0.8\linewidth]{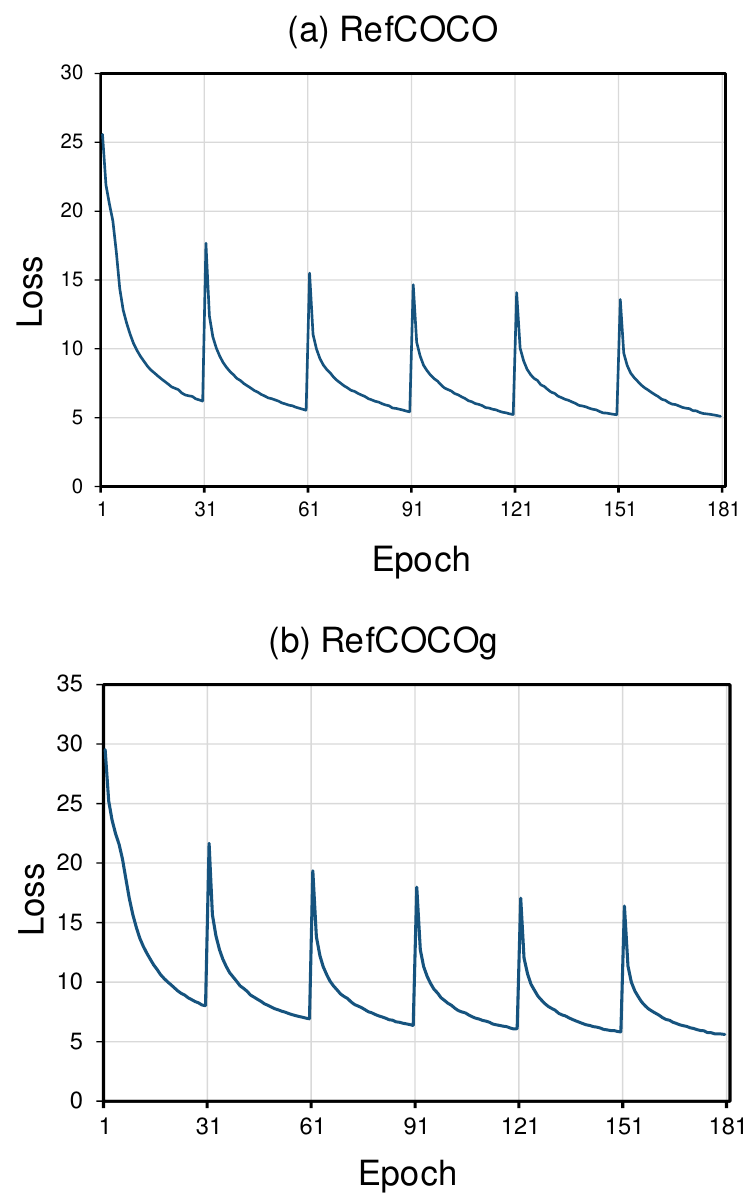}
\vspace{-4mm}
\caption{Training loss of SiRi mechanism on (a)~RefCOCO and (b)~RefCOCOg using constant learning rate. }
\label{fig:training loss}
\end{wrapfigure}

For the error cases, we found the network usually fails if the referred object is obscured or occluded, \eg, in ``bread closer to bowl", the target object is occluded.
Another common error case is that the referring expression is based on the text content on the object, \eg, ``happy birthday cake''.

Fig.~\ref{fig:training loss} depicts the loss curve of the training process using SiRi mechanism. The retraining period is set as 30 epochs. As can be seen from the loss curves, the model reaches a better local minimum after each retraining progress. It verifies our motivation that a better initialized encoder for vision-language perception usually helps the model converge to a better local minimum.

\subsection{Comparison with Large Pre-training}
We report SiRi \textit{with} large pre-training in the table below. We can see that SiRi could further improve even when large-scale pre-training has provided superior initialization for the whole model.

\begin{table}[h]
\centering
\setlength{\tabcolsep}{3mm}{
\begin{tabular}{c|ccc|ccc} 
\toprule
\multirow{2}{*}{Model}                                    & \multicolumn{3}{c|}{RefCOCO} & \multicolumn{3}{c}{RefCOCO+}  \\
                                                          & val   & testA & testB        & val   & testA & testB         \\ 
\hline\hline
\begin{tabular}[c]{@{}c@{}}MDETR (pretrained)\end{tabular} & 86.75 & 89.58 & 81.41        & 79.52 & 84.09 & 70.62         \\
+SiRi                                                     & 87.24 & 89.57 & 81.83        & 79.77 & 84.28 & 70.98         \\
\bottomrule
\end{tabular}}
\setlength{\abovecaptionskip}{0.1cm}
\caption{Experiment results of MDETR (large pretraining) with SiRi on RefCOCO and RefCOCO+.}
\label{tab:pretrain}
\end{table}

\end{document}